\documentclass{article}

\usepackage[preprint,nonatbib]{confstyle}

\usepackage[utf8]{inputenc} 
\usepackage[T1]{fontenc}    
\usepackage{hyperref}       
\usepackage{url}            
\usepackage{booktabs}       
\usepackage{amsfonts}       
\usepackage{nicefrac}       
\usepackage{microtype}      
\usepackage{xcolor}         
\usepackage{graphicx}  
\usepackage{setspace}
\usepackage{multirow}
\usepackage{hyperref}

\usepackage{mathptmx} 
\usepackage{times} 
\usepackage{amssymb}  

\usepackage{listings} 

\usepackage{algorithm}  
\usepackage{algorithmic}
\usepackage{amsmath}

\usepackage{balance}
\usepackage{adjustbox}
\usepackage{tabularx}

\title{\Large BotDirector: Robot Storytelling Across the Symmetrical Reality with Multi-modal Interactions}

\author{
    Zhe Sun$^{1}$ \quad
    Meng Wang$^{1}$ \quad
    Lei Wang$^{1}$ \quad
    Yuxi Wang$^{2}$ \quad
    Wanxin Li$^{2}$ \\[1ex]
    \textbf{Yujia Peng$^{1,2}$ \quad
    Zhenliang Zhang$^{1}$} \quad \thanks{e-mail: zlzhang@bigai.ai} \\
    \scriptsize
    $^{1}$State Key Laboratory of General Artificial Intelligence, BIGAI, Beijing, China \\
    \scriptsize
    $^{2}$Peking University, Beijing, China
}

\makeatletter
\def\thanks#1{\protected@xdef\@thanks{\@thanks
        \protect\footnotetext{#1}}}
\makeatother

\begin{document}

\maketitle

\begin{figure}[htbp]
  \centering
  \includegraphics[width=\linewidth]{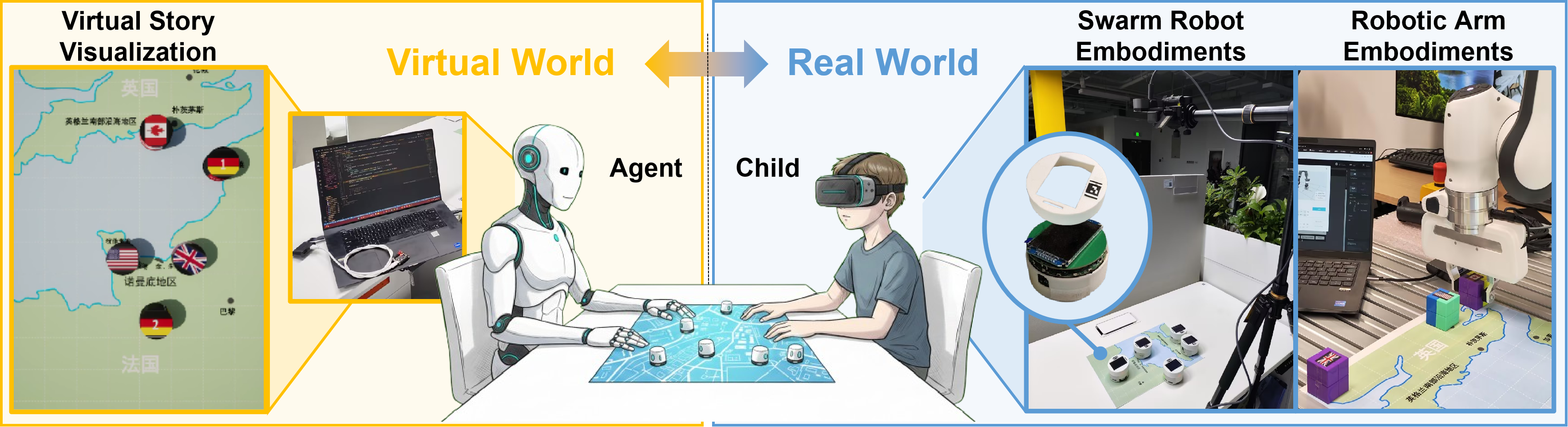}
  \caption{BotDirector, a tangible interactive system that helps children to quickly create a desk-top drama with their daily items (e.g.: personal toys), swarm robots (e.g.: desktop robots with screens), and robotic arms (e.g.: Franka robotic arm).}
  \label{fig:teaser}
\end{figure}

\begin{abstract}
Robot storytelling offers a unique blend of technological innovation and creative expression that engages children in unprecedented ways. However, the technical aspects are often too complicated for children. We propose an interactive system that facilitates robot storytelling with tangible and natural language interactions. Children arrange the playground with their own stuff and create narratives with an LLM agent. The created narratives are transformed into a motion sequence based on the map and characters, and the motions are executed by self-navigating swarm robots. This system enhances robot storytelling with flexible scenarios, enabling young children to create robot dramas with everyday objects.
\end{abstract}

\section{Introduction}

Robot storytelling has emerged as a compelling domain that offers a unique blend of technological innovation and creative expression\cite{jouen2025once, zhang2024mathemyths}. By integrating robotics into narrative practices\cite{liang2023robot}, this medium engages children in unprecedented ways, fostering both cognitive development and imaginative play. However, a significant barrier remains: the technical complexity inherent in programming and robot control is often too demanding for young children, limiting their role to passive observers rather than active creators.

To bridge this gap, we propose an interactive system designed to facilitate robot storytelling through intuitive tangible and natural language interactions for children between 5 to 8 years old. Instead of navigating complex interfaces, children can simply arrange a ``playground'' using their own everyday objects. By interacting with an Large Language Model (LLM) agent, children can collaboratively create narratives that are automatically transformed into precise motion sequences. These motions are then executed by robots that interpret the physical map and character roles defined by the user. We raised two research questions: 

\begin{itemize}
    \item \textbf{RQ1}: How can the integration of natural language interaction via LLMs and tangible object arrangement effectively lower the technical barrier for young children to design complex robot narratives?
    \item \textbf{RQ2}: In what ways does the use of physical embodiments (e.g.: swarm robots, robotic arms) and everyday objects (e.g.: toys, LEGO bricks) enhance the creative flexibility and narrative engagement of children's storytelling?
\end{itemize}

By leveraging the synergy between natural language processing and robotic embodiments, this system enhances robot storytelling with flexible scenarios. Ultimately, it empowers young children to transform their domestic environments into stages, enabling them to create dramas in collaboration with robots with ease and autonomy.

\section{Method}
\subsection{Interactive Design}
The BotDirector system comprises three stages: (1) knowledge generation, (2) script generation, and (3) interactive play. In the knowledge generation stage, children initiate the process by proposing a topic of interest and engaging in a dialogue with the system to iteratively expand and refine the theme into a cohesive narrative. For example, a child may select a historical theme and gradually abstract the events of a famous war into a structured story through conversational interaction. During the script generation stage, the planning module derives a formal script from the narrative and generates a virtual environment with digital embodied agents corresponding to the story’s characters. The interactive play stage then follows, where the system executes the script through these agents while the child provides real-time feedback using natural language. This feedback mechanism allows children to modify both the narrative content, such as redirecting a robot's movement from one coordinate to another, and the physical representation of characters, which can be selected from swarm robots and bricks that the BotDirector system provides, or personal items like toys and LEGO bricks. These modifications are performed interactively, enabling children to interrupt the performance and suggest revisions without any programming knowledge, utilizing only natural language and tangible interfaces. By facilitating this hands-on practice in theatrical creation, the design aims to increase student engagement, stimulate creativity, and enhance the retention and understanding of the specific knowledge points embedded within the stories.

\subsection{Virtual-Physical Synergy}
The virtual environment is primarily utilized in Stage 2, where BotDirector presents generated script drafts to children to accelerate the efficiency of narrative generation and iteration, whereas the physical environment is employed in Stage 3 to allow children to observe BotDirector controlling swarm robots or robotic arms to animate personal items and to interact with the system by moving these physical embodiments to facilitate script modifications. 

Leveraging the theory of symmetrical reality\cite{zhang2024emergence}, we align the virtual and physical environments, utilizing the virtual domain in Stage 3 to guide children during the assignment of physical embodiments to narrative characters. By synchronizing the virtual stage map with the physical workspace, the system highlights specific characters in the virtual environment, prompting the child to place their selected physical object at the corresponding coordinates in the physical space to complete the role assignment. These virtual-physical pairings subsequently serve as spatial anchors that maintain the alignment between the virtual and physical domains throughout the interaction.

\section{Implementation}
\subsection{Interactive Story and Script Completion}
Within the BotDirector system, we integrated GPT-4 as the core reasoning and decision-making component to facilitate natural language parsing, narrative expansion, and script generation. Specifically, Stage 1 utilizes three prompting modes: generating a core narrative based on a user-defined theme, answering user inquiries to augment the story, and posing questions to the user to stimulate further creative thought, with the latter two modes iterating until the user considers the narrative sufficiently complete.

In Stage 2, BotDirector automatically generates a script draft through a five-step pipeline that includes extracting character counts and names from the narrative, generating character portraits, creating capsule-based virtual embodiments using these portraits as textures while mapping character identities to robot identifiers, selecting an appropriate stage from a predefined map library, and ultimately synthesizing a structured JSON script that integrates the story, character-robot pairs, and spatial parameters.

The script generated in Stage 2 is like \{robot\_id, initial coordinates, target coordinates, character dialogue, narrator voiceover\}, with all spatial data defined within the coordinate system of the virtual map. To facilitate the physical execution in Stage 3, we developed a planning module for the BotDirector system that performs a spatial transformation to map these virtual coordinates onto the corresponding coordinate system of the physical environment. The moving trajectory of robots is calculated by the A* algorithm. 

Note that there are two kinds of interactions between children and the BotDirector: natural language interaction and tangible interaction. Natural language interaction is implemented as follows: children express their thoughts verbally, which the BotDirector interprets via a Speech-to-Text (STT) module to generate feedback; this feedback is then played back to the children through a Text-to-Speech (TTS) module. Physical interaction is realized by allowing children to directly move objects corresponding to specific characters if they are dissatisfied with their placement; the BotDirector then tracks and records the updated character positions using an object detection module.

\subsection{Reflective Iteration and Error Correction}

In Stage 3, users interactively engage with BotDirector to reflect upon and optimize the script draft generated in Stage 2. As BotDirector repeatedly performs the script, users can interrupt at any point via natural language to request modifications to the narrative or execution. A continuous speech-to-text module captures these verbal inputs, converting them into text which is then processed by a parser to populate pre-designed prompt templates. This prompt, incorporating the user’s specific revision requirements, is fed into the LLM, which subsequently generates an updated and refined script.

To compensate for the cumulative movement errors in the IMU-based swarm robots, and to provide the global localization required for the robotic arm to manipulate personal items, we deployed a monocular overhead RGB camera to capture a bird's-eye view of the physical stage. The boundaries of the stage are demarcated by AprilTags placed at the four corners, ensuring a stable spatial reference frame. Furthermore, supplementary AprilTags are affixed to each swarm robot and the user’s personal items to facilitate real-time coordinate tracking. These dynamic positional data are integrated into the motion planning module to continuously calibrate and adjust the moving trajectories during the performance.

\subsection{Embodiment Collaboration in Symmetrical Reality}

We pre-designed a set of paired maps for both the virtual and physical environments, where each pair features identical geometries and shares a synchronized coordinate system, serving as a fundamental component for cross-reality alignment. In addition to these spatial frameworks, Stage 3 allows users to assign physical embodiments to specific characters as described in Section 3.2; when a user specifies an embodiment, the virtual character is paired with that physical object, whereas the BotDirector system automatically assigns an embodiment in the absence of user input. These virtual-physical embodiment pairings constitute another critical element in achieving alignment between the virtual and physical domains.

\section{Conclusion}
In this poster, we present a system named BotDirector for child-robot collaborative theatrical performance, encompassing the design of the system architecture, interaction workflows, and the construction of a functional prototype. Moving forward, we plan to conduct user studies with children to evaluate the usability and effectiveness of the proposed system in addressing our two primary research questions. Furthermore, we intend to explore designs that more deeply integrate virtual and physical spaces, specifically by leveraging virtual environments to extend the performance area when physical space is insufficient for the play. This approach includes generating multiple auxiliary virtual stages that synchronize with the physical narrative, allowing virtual characters to enact scenes that exceed the spatial capacity of the physical stage.


\begin{thebibliography}{1}

\bibitem{jouen2025once}
A.-L. Jouen, R.~Matsunaka, and K.~Hiraki.
\newblock Once upon a time… acquisition of second language vocabulary through robotic storytelling in classroom settings.
\newblock {\em International Journal of Social Robotics}, pages 1--34, 2025.

\bibitem{liang2023robot}
J.-C. Liang and G.-J. Hwang.
\newblock A robot-based digital storytelling approach to enhancing efl learners’ multimodal storytelling ability and narrative engagement.
\newblock {\em Computers \& Education}, 201:104827, 2023.

\bibitem{zhang2024mathemyths}
C.~Zhang, X.~Liu, K.~Ziska, S.~Jeon, C.-L. Yu, and Y.~Xu.
\newblock Mathemyths: Leveraging large language models to teach mathematical language through child-ai co-creative storytelling.
\newblock {\em arXiv preprint arXiv:2402.01927}, 2024.

\bibitem{zhang2024emergence}
Z.~Zhang, Z.~Zhang, Z.~Jiao, Y.~Su, H.~Liu, W.~Wang, and S.-C. Zhu.
\newblock On the emergence of symmetrical reality.
\newblock In {\em Proceedings of the IEEE Conference Virtual Reality and 3D User Interfaces (VR)}, pages 639--649. IEEE, 2024.

\end{thebibliography}


\end{document}